\documentclass[10pt,twocolumn,letterpaper]{article}

\usepackage[pagenumbers]{cvpr} 

\definecolor{HER2}{HTML}{EE8636} 
\definecolor{BASAL}{HTML}{3B75B0} 
\definecolor{Normal}{HTML}{8E69B8} 
\definecolor{LumB}{HTML}{D62627} 
\definecolor{LumA}{HTML}{519D3F} 

\usepackage{pifont}
\usepackage{multirow}
\usepackage{arydshln}
\usepackage{xcolor}
\usepackage{mathrsfs}

\definecolor{cvprblue}{rgb}{0.21,0.49,0.74}
\usepackage[pagebackref,breaklinks,colorlinks,allcolors=cvprblue]{hyperref}
\usepackage[accsupp]{axessibility} 

\def\paperID{6863} 
\def\confName{CVPR}
\def\confYear{2026}

\title{Cell-Type Prototype-Informed Neural Network \\
for Gene Expression Estimation from Pathology Images}

\author{
Kazuya Nishimura$^{1, 3}$ \quad Ryoma Bise$^{2}$ \quad Shinnosuke Matsuo$^{2}$ \quad Haruka Hirose$^{3}$ \quad Yasuhiro Kojima$^{3}$\\
$^{1}$ The University of Osaka, Japan \quad $^{2}$Kyushu University \quad $^{3}$National Cancer Center Japan\\
{\tt\small k.nishimura.d3c@osaka-u.ac.jp}
}

\begin{document}
\maketitle
\begin{abstract}
Estimating slide- and patch-level gene expression profiles from pathology images enables rapid and low-cost molecular analysis with broad clinical impact. 
Despite strong results, existing approaches treat gene expression as a mere slide- or spot-level signal and do not incorporate the fact that the measured expression arises from the aggregation of underlying cell-level expression.
To explicitly introduce this missing cell-resolved guidance, we propose a Cell-type Prototype-informed Neural Network (CPNN) that leverages publicly available single-cell RNA-sequencing datasets. Since single-cell measurements are noisy and not paired with histology images, we first estimate cell-type prototypes—mean expression profiles that capture stable gene–gene co-variation patterns. CPNN then learns cell-type compositional weights directly from images and models the relationship between prototypes and observed bulk or spatial expression, providing a biologically grounded and structurally regularized prediction framework.
We evaluate CPNN on three slide-level datasets and three patch-level spatial transcriptomics datasets. Across all settings, CPNN achieves the highest performance in terms of Spearman correlation. Moreover, by visualizing the inferred compositional weights, our framework provides interpretable insights into which cell types drive the predicted expression.
Code is publicly available at \url{https://github.com/naivete5656/CPNN}.
\end{abstract}

\section{Introduction}
\label{sec:intro}
Estimating gene expression directly from pathology whole-slide images (WSIs) has emerged as a promising alternative to RNA sequencing, enabling large-scale molecular profiling at a fraction of the cost. This capability supports molecular subtype classification, personalized treatment stratification, and investigations into how histological morphology reflects underlying gene expression \cite{graziani2022attention, Alsaafin2022LearningTP, zheng2024digital, csenbabaouglu2024mosby, pizurica2024digital}.

\begin{figure}
    \centering
    \includegraphics[width=0.98\linewidth]{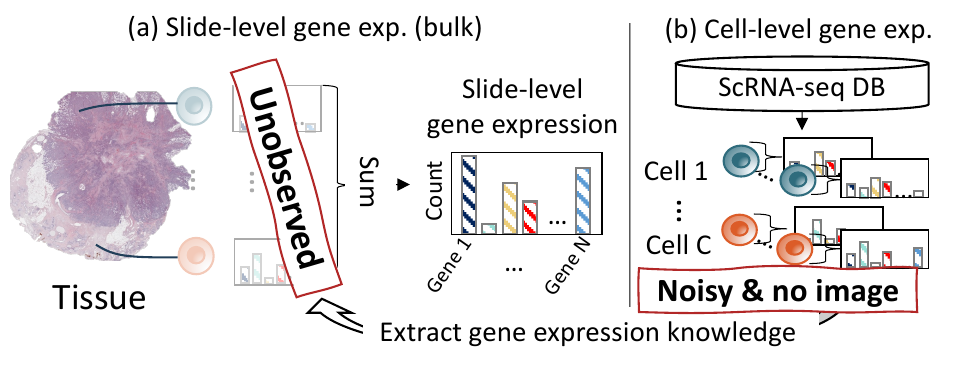}
    \vspace{-4mm}
    \caption{(a) Gene expression is observed together with histology image as the sum of unobserved cell-level expressions. Expression of individual cells is not directly measurable. (b) Cell-level gene expression profiles are obtained from a public single-cell RNA sequencing (scRNA-seq) database, which contains noisy measurements without corresponding image information.
    \vspace{-5mm}
    }
    \label{fig:intro}
\end{figure}

Existing methods infer gene expression at the population level rather than at individual cells. Current approaches can be broadly categorized into two major settings.
1) Slide-level estimation which targets bulk transcriptomics, predicts expression aggregated over an entire tissue section. In this setting, multiple-instance learning architectures have been proposed to capture global tissue representations \cite{graziani2022attention, schmauch2020deep, Alsaafin2022LearningTP, pizurica2024digital, csenbabaouglu2024mosby}.
2) Patch-level estimation which targets spatial transcriptomics, predicts expression in localized regions containing multiple cells. 
Models in this setting leverage both local and contextual features around each spatial spot using transformer or graph-convolution architectures \cite{pang2021leveraging, he2020integrating, zeng2022spatial, yang2023exemplar, chung2024accurate}.

While existing methods learn effective image features for the expression estimation, they do not explicitly model the underlying data-generation process in which the observed expression arises from the aggregation of cell-level signals. As shown in Figure \ref{fig:intro}(a), gene expression originates at the level of individual cells, but only the population-level aggregate is observed in slide- or patch-level measurements. Consequently, current approaches rely solely on these aggregated signals and cannot incorporate cell-resolved biological structure.

To address the lack of cell-level information, one natural direction is to leverage single-cell RNA-sequencing (scRNA-seq) data, which provide cell-level expression profiles for similar biological tissues. Such datasets are publicly available and often annotated with cell types  in databases such as GEO \cite{barrett2012ncbi}, SCPortalen \cite{Abugessaisa2017SCPortalenHA}, and the Single Cell Portal \cite{Tarhan2023SingleCP} (Figure \ref{fig:intro}(b)). However, they are noisy, affected by batch effects, and lack matched histopathology images, making their direct use for WSI-based gene expression estimation challenging.

\begin{figure}
    \centering
    \includegraphics[width=0.9\linewidth]{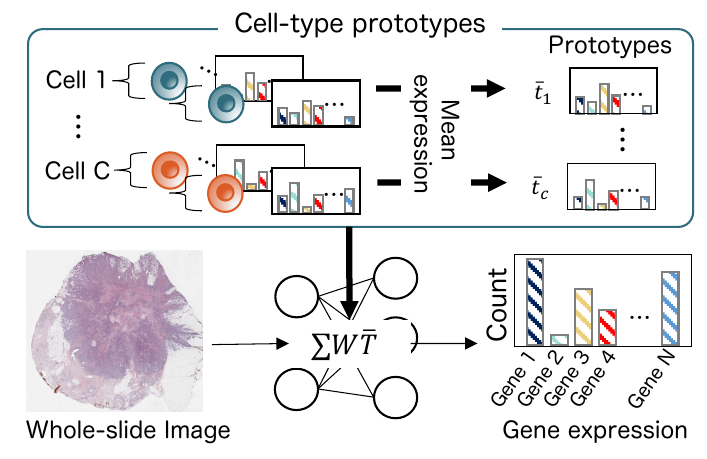}
    \caption{Illustration of concept of Cell-type Prototype-informed Neural Network (CPNN). We estimate gene expression based on cell-type prototypes by modeling the relationship between observed gene expression and cell-level gene expression.}
    \label{fig:intro2}
    \vspace{-5mm}
\end{figure}

To leverage single-cell information despite its noise and lack of correspondence with WSIs, we introduce a Cell-type Prototype-informed Neural Network (CPNN). The method assumes that slide- and patch-level gene expression can be represented as mixtures of cell-type gene expression profiles. CPNN estimates cell-type prototypes $\bar{T}$ and compositional weights $W$, as illustrated in Figure~\ref{fig:intro2}. The predefined prototypes, derived from cell-level datasets, encode stable gene–gene covariation patterns for each cell type and serve as informative priors for estimation. During training, they are further refined to mitigate noise and reduce the modality gap between cell- and slide-level measurements.

We evaluate CPNN on three slide-level datasets (BRCA, KIRC, LUAD) and three patch-level spatial transcriptomics datasets (CSCC, Her2st, STNet). 
For patch-level prediction, CPNN can be integrated with existing architectures to fully leverage their feature extraction capabilities. 
Across all datasets, CPNN consistently achieves the highest Spearman correlation. 
In addition, visualizing the inferred compositional weights reveals which cell types drive the predictions, providing interpretable insights into the model's decision process.

Our contributions are summarized as follows:
\begin{itemize}
    \item A problem formulation for gene expression estimation from WSIs that leverages additional cell-level gene expression datasets retrieved from single-cell databases.
    \item A Cell-type Prototype-informed Neural Network (CPNN) that integrates cell-type prototypes and their compositional weights to predict gene expression.
    \item Comprehensive experiments on publicly available slide- and patch-level datasets demonstrating the superior performance and generalization ability of the our method.
\end{itemize}

\section{Related work}

\noindent
{\bf Slide-Level Gene Expression Estimation.}
Slide-level gene expression estimation from whole slide images treated as a multiple-instance regression (AbReg) \cite{graziani2022attention}. To reduce the overfitting problem, Schmauch et al. have reduced instances by clustering (HE2RNA) \cite{schmauch2020deep}. Alsafin et al. have introduced a transformer (tRNAformer) \cite{Alsaafin2022LearningTP}, and Pizurica et al. have introduced lineage operation (SEQUOIA VIS) \cite{pizurica2024digital} for gene expression estimation. {\c{S}}enbabao{\u{g}}lu et al. utilized patch-level representations derived from self-supervised learning, combined with an MLP, to estimate gene expression (MOSBY) \cite{csenbabaouglu2024mosby}.
However, the effectiveness of these modules has not been rigorously compared with the MIL methods on fair benchmarking.

\noindent
{\bf Patch-Level Gene Expression Estimation.}
The emergence of spatial transcriptomics has stimulated the development of methods that infer gene expression directly from localized image patches \cite{he2020integrating, zeng2022spatial, jia2023thitogene, ganguly2025merge, pang2021leveraging, chung2024accurate, xie2023spatially, yang2023exemplar, dang2025hage}.
Following the pioneering convolutional neural network (CNN)–based approach STNet introduced by \cite{he2020integrating}, researchers have extended this line of work by incorporating graph neural networks (GNNs) \cite{zeng2022spatial, jia2023thitogene, ganguly2025merge} and transformer architectures such as TRIPLEX \cite{chung2024accurate} to better capture spatial and contextual relationships among tissue regions.
Another line of research aims to utilize retrieved gene expression to estimate expression profiles, as demonstrated by models such as BLEEP \cite{xie2023spatially} and EGN \cite{yang2023exemplar}, as well as HAGE \cite{dang2025hage}.
While these retrieval-guided methods can help the model produce predictions that better approximate real observations, accurate alignment between image and gene expression remains difficult.
These methods can not resolve the challenges of lack of cell-level information. 

In contrast, our method complements the lack of cell-level information by introducing cell-level gene expression retrieved from database.



\noindent
{\bf Single-Cell Deconvolution.}
Single-cell deconvolution, which infers the cell-type composition of slide- and patch-level expression using a single-cell dataset, has been proposed \cite{wang2019bulk, gynter2023deconv, chu2022cell, kojima2024single}. 
These methods model the relationship between the mean gene expression for each cell type and the gene expression and estimate the cellular composition that reconstructs the slide-level gene expression.

This work, for the first time, leverages the relationship between target gene expression and cell-type prototypes to estimate slide- or patch-level gene expression, differing fundamentally from deconvolution-based approaches.



\noindent
{\bf Physics-Informed Neural Network.}
A physics-informed neural network is a technique that integrates physical laws and data-driven representation of neural network \cite{toscano2024pinns}. The technique is widely used for various application fields such as imaging \cite{kamali2023elasticity}, super-resolution \cite{yasuda2022super}, and biomedicine \cite{jo2024density}. These methods enable the network to estimate regularized estimation via loss functions or model architectures with a small amount of training data. 
Our approach is similar to PINN in that it integrates gene expression with cell-level guidance based on physical relationships.

\noindent 
{\bf Multiple-Instance Learning.}
Multiple-instance learning is the task that estimates bag-level objectives from multiple instances ({\it i.e.,} a set of images) \cite{wang2018revisiting, ilse2018attention, xiang2023exploring}.
Research trends focus on efficiently aggregating features from multiple instances to generate bag features. Early methods relied on traditional approaches such as max pooling and mean pooling (Max and Mean) \cite{wang2018revisiting}. 
Attention mechanisms (AbMIL) \cite{ilse2018attention} and transformers (ILRA) \cite{xiang2023exploring} have been introduced for more smart instance feature aggregation. 
A recent trend focuses on leveraging state space model to capture global contextual features, enabling more comprehensive representations (S4MIL, MambaMIL, SDMamba) \cite{fillioux2023structured, yang2024mambamil, zhang20252dmamba}.

\begin{table}[t]
    \caption{
        Comparison of bulk, single-cell, and spatial transcriptomics techniques.
    }
    \label{tab:transcriptomics}
    \centering
    \begin{tabular}{lccc}
    \hline
    \textbf{Techniques} & \textbf{Bulk} & \textbf{Single-cell} & \textbf{Spatial} \\ \hline
    With image
       & \checkmark
       & \ding{55} 
       & \checkmark \\
    Exp. resolution
       & Tissue 
       & Single-cell 
       & Spot \\
    Number of samples
       & High 
       & Moderate 
       & Low \\
    Cost per sample
       & Low 
       & Medium 
       & High \\
    \hline
    \end{tabular}
    \vspace{-5mm}
\end{table}

\noindent
{\bf Sequencing Techniques of RNA.}
RNA sequencing provides a comprehensive profile of RNA transcripts ({\it i.e.,} gene expression) and is essential for examining cellular function. 
There are some techniques, including bulk, single-cell, and spatial transcriptomics, each with distinct trade-offs in data scale, resolution, cost, and complexity. Table \ref{tab:transcriptomics} shows a summary of the trade-off of each modality.
\begin{itemize}
    \item Bulk transcriptomics ({\it i.e.,} slide-level gene expression) \cite{mortazavi2008mapping} provides the gene expression level across the cells in each biological specimen. It is relatively low cost, noise is less than other techniques, and there is a large-scale database that contains paired whole slide images and gene expression. However, it cannot capture cell-level activity. 
    \item Single-cell transcriptomics \cite{jovic2022single} capture the gene expression of individual cells. However, it lacks corresponding imaging data, loses spatial context within tissues, and is more susceptible to noise.
    \item Spatial transcriptomic ({\it i.e.,} patch-level gene expression) \cite{marx2021method} was developed to capture gene expression with spatial context and image. Although it can capture spatial context, spatial transcriptomics remains technically demanding and generally more costly.
    
\end{itemize}

\begin{figure*}[t]
    \centering
    \includegraphics[width=0.95\linewidth]{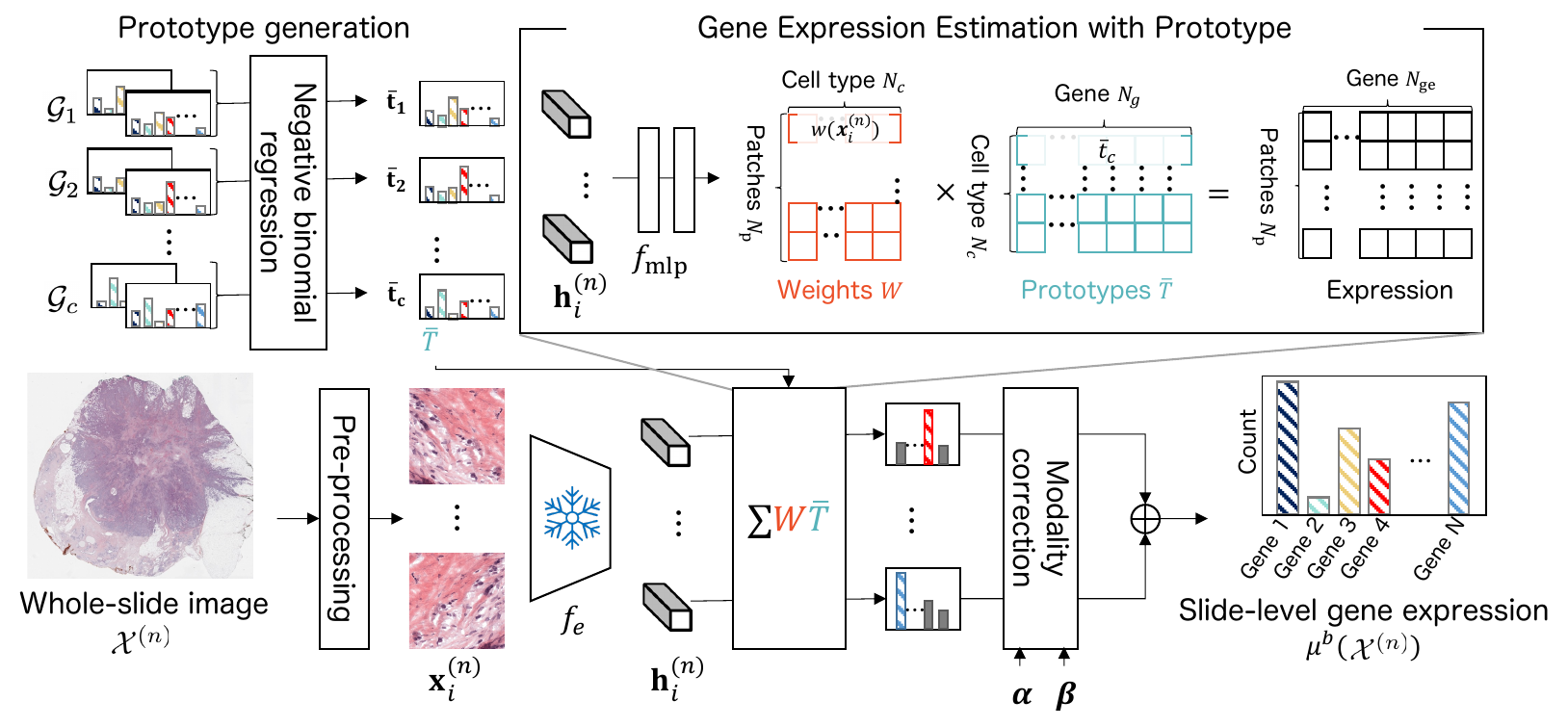}
    \caption{Overview of our method. Gene expression is estimated based on cell-type prototype $\bar{T}$ and weight $w (\mathbf{x}_i^{(n)} )$. The prototype supports the integration of covariance of gene expression among cells.}
    \label{fig:overview}
\end{figure*}


\section{Cell-Type Prototype-Informed Neural Network}
\label{sec:formatting}

In this section, we introduce our approach for modeling slide-level gene expression (Section~\ref{sec:slide-level}).
The same framework can also be applied to patch-level estimation by replacing the whole-slide image with patch images, as described in Section~\ref{sec:patch-level}.

\subsection{Slide-Level Estimation}
\label{sec:slide-level}
The goal of slide-level gene expression estimation is to estimate slide-level gene expression profiles from whole-slide pathology images.
Given the $n$-th pathology whole-slide image $\mathcal{X}^{(n)} = \left\{ \mathbf{x}_i^{(n)} \right\}_{i=1}^{N_{\mathrm{p}}}$, where each $\mathbf{x}_i^{(n)}$ is a patch image, we aim to predict the slide-level gene expression vector $\mathbf{e}^{\mathrm{b} (n)} \in \mathbb{R}^{N_{\mathrm{ge}}}$, where each element corresponds to the expression count of a gene. $N_{\mathrm{p}}$ and $N_{\mathrm{ge}}$ denote the number of patches and genes.
To regularize training, we incorporate cell-level expression dataset (single-cell RNA-seq.) from public sources, denoted as $\mathcal{G}_c = \left\{ \mathbf{e}^{\mathrm{sc} (k)}_c \right\}^{N_{\mathrm{sc}}}_{k=1}$,  where $\mathbf{e}^{\mathrm{sc} (k)}_c$ is a $k$-th single-cell-level gene expression of the cell type $c$. $N_{\mathrm{sc}}$ is the number of single-cell samples of $c$.

An overview of the proposed method is shown in Figure \ref{fig:overview}. 
We assume that the expression level of each patch is an aggregation of cell-level expression across the patch and define slide-level expression using the gene expression prototype $\bar{T}$ obtained from the obtained cell-level dataset and its compositional weight (Section \ref{sec:formulation}).
First, we generate the gene expression prototype $\bar{T}$ from the cell-level dataset (Section \ref{sec:prototype}). The compositional weight $w$ is estimated from the patch cropped from the slide $\mathcal{X}^{(n)}$ (Section \ref{sec:proportion}).
We optimize our model with a negative log-likelihood for a slide-level gene expression and regularization (Section \ref{sec:optimization}).

\subsubsection{Formulation of Slide-Level Expression}
\label{sec:formulation}
We model the mean gene expression $\mu^\mathrm{b}_g \left(\mathcal{X}^{(n)} \right)$ as an aggregation of cell-type–specific prototypes with composition weights for the prototypes:
\vspace{-2mm}
\begin{equation} \label{eq:model}
    \mu^{\mathrm{b}}_g \left(\mathcal{X}^{(n)} \right) = \underbrace{\alpha_g}_{\text{Tech. effect}}  \sum_i \sum_c \overbrace{w\left(\mathbf{x}_i^{(n)} \right)_{c}}^{\text{Comp. weights}}  \underbrace{\bar{T}_{c,g}}_{\text{Prototypes}} + \overbrace{\beta_g}^{\text{Tech. effect}},
\end{equation}
where $\alpha_g$ represents a gene-specific scaling factor, which is expected to reflect the technical capturing rate of the corresponding gene, and $\beta_g$ is a shift parameter due to the gap of modality for the gene $g$. 


The total count of genes in the sample $l^{(n)}$ depends on factors that are difficult to estimate from the slide, such as tissue thickness. 
Therefore, our model estimates mean gene expression that is independent of the total count of the sample and achieves robustness for these factors.

In order to make our model estimates independent of the total counts across genes, 
We divid the $\mathbf{\mu}^{\mathrm{b}} \left(\mathcal{X}^{(n)} \right)$ by the summation of it across genes and multiplied the actual total count of the sample $n$, which is $l^{(n),}$.
\begin{equation} \label{eq:model2}
    \bar{\mu}^{\mathrm{b}}_g \left(\mathcal{X}^{(n)}\right) = l^{(n)} \frac{\mu_g^{\mathrm{b}} \left(\mathcal{X}^{(n)} \right)}{\sum_g \mu_g^{\mathrm{b}}\left(\mathcal{X}^{(n)}\right)}.
\end{equation}
Similar to previous studies \cite{lopez2018deep}, we assume that the observed slide-level gene expression $\mathbf{e}^{\mathrm{b} (n)}$ follows a Negative Binomial distribution. 
The conditional probability of $e^{\mathrm{b} (n)}$ given the slide $\mathcal{X}^{(n)}$  can be formulated as follows:
\begin{align}
    \Pr \left(\mathbf{e}^{\mathrm{b} (n)} | \mathcal{X}^{(n)} \right) = \prod_g \Pr \left(e^{\mathrm{b} (n)}_g|\mathcal{X}^{(n)} \right), \\
    \label{eq:dist}
    \Pr \left(e^{\mathrm{b} (n)}_g|\mathcal{X}^{(n)} \right) = \mathrm{NB}\left(\bar{\mu}^{\mathrm{b}}_g \left(\mathcal{X}^{(n)} \right), \theta^{\mathrm{b}}_g \right),
\end{align} 
where $\mathrm{NB}$ denotes the Negative Binomial distribution, $\bar{\mu}^{\mathrm{b}}_g(\mathcal{X}^{(n)})$ represents the mean expression level of gene $g$ computed using Eq. \ref{eq:model2}, and $\theta^{\mathrm{b}}_g$ is the dispersion parameter for gene $g$.

\subsubsection{Batch-Agnostic Prototype Generation}
\label{sec:prototype}
We first estimate gene expression prototypes $\bar{T}$ from the cell-level dataset $\mathcal{G}_c$ using negative binomial regression. 
Since single-cell data tend to contain an experimental batch effect ({\it i.e.,} a gap or shift due to a technical factor) between observed samples, we generate prototypes to estimate normalized gene expression using the negative binomial regression model \cite{kleshchevnikov2022cell2location}. 

We assume that the observed cell-level gene expression count follows a Negative Binomial distribution.
\begin{equation}
   e^{\mathrm{sc}(k)}_{c,g}  \sim \mathrm{NB} \left( \mu^{\mathrm{sc}}_{c, g}, \theta^{\mathrm{sc}}_{c, g} \right).
\end{equation}
We model $\mathbf{\mu}^{\mathrm{sc}}_c$  using technical effects and the mean expression for each cell type $t^{\mathrm{sc}_c}$.
\begin{equation}
    \mu^{\mathrm{sc}}_{c,g} = \left(t_{c,g}  +  b_d \right) s_d,
\end{equation}
where $s_d$ is a scaling parameter between the experimental condition $d$, and $b_d$ is an additive gene-wise background shift of $d$.
The prototype for each cell type $\mathbf{t}_{c}\in \mathbb{R}^{N_{\mathrm{ge}}}$ is obtained by fitting observed gene expression with the regression model. 
The estimated prototype is normalized by $\mathbf{\bar{t}}_{c} = \frac{\mathbf{t}_{c}}{\sum_g \mathbf{t}_{c}}$, and  $\bar{T} = \left[\mathbf{\bar{t}}_{1 \top}, ..., \mathbf{\bar{t}}_{N_{\mathrm{sc}} \top} \right]^{\top}$ is obtained.


\subsubsection{Gene Expression Estimation with Prototype} \label{sec:proportion}
Next, we estimate the compositional weight for prototypes from the whole-slide image $\mathcal{X}^{(n)}$.
Given the slide $\mathcal{X}^{(n)}$, we extract image features $\mathbf{h}_i^{(n)}$ for each patches $\mathbf{x}_i^{(n)}$ that is cropped from $\mathcal{X}^{(n)}$.
We follow the pre-processing paradigm of MIL \cite{lu2021data}, which crops small patches from tissue regions and extracts the feature by an off-the-shelf image encoder $f_{\mathrm{e}}$. 
The image feature $\mathbf{h}_i^{(n)} = f_{\mathrm{e}}(\mathbf{x}_i^{(n)} )$ is extracted from each patch $\mathbf{x}_i^{(n)}$.
The weight is estimated from the extracted features $\mathbf{h}_{i}^{(n)}$ by the two linear layers $f_{\mathrm{mlp}}$. 
Cell-type proportion for patch is obtained $w(x_i^{(n)}) = \mathrm{softmax} \left(f_{\mathrm{mlp}}\left(\mathbf{h}_i^{(n)}\right) \right)$. 
As shown in Figure \ref{fig:overview} of CPNN, the gene expression estimation is calculated by the dot product of the compositional weight and cell-type prototype $\bar{T}$. Gene expression for each patch is computed. 

Cell-level and slide-level expression often exhibit gene-specific shifts and biases due to differences in measurement technology between these two modalities. To address these discrepancies, we introduce 
$\boldsymbol{\alpha}$ and $\boldsymbol{\beta}$ as scale and bias corrections. As shown in Equation \ref{eq:model}, we derive bulk gene expression by applying these correction parameters to the gene expression obtained from the matrix product.
Finally, the mean parameter $\bar{\mu}^{\mathrm{b}} \left(\mathcal{X}^{(n)} \right)$ of Negative Binomial Distribution is obtained by Equation \ref{eq:model} and \ref{eq:model2}.

\subsubsection{Optimization}
\label{sec:optimization}
The main objective of our optimization is to reconstruct gene expression data.
If only reconstruction loss is used for optimization, there is no grounding for prototype and proportions, and the optimization may lose the cell-level information.
To avoid this issue, We impose a soft consistency constraint with the deconvolution results \cite{kleshchevnikov2022cell2location}. 
Our optimization updates $\boldsymbol{\theta}^{\mathrm{b}}$, $\boldsymbol{\alpha}$, $\boldsymbol{\beta}$, $\bar{T}$, and $w (\mathbf{x}_i^{(n)} )$. 
To update $w (\mathbf{x}_i^{(n)} )$, the parameter of neural network $f_{\mathrm{mlp}}$ is updated.

The total loss function of our method is as follows:
\vspace{-1mm}
\begin{align}
    L_{\mathrm{total}} = L_{\mathrm{NB}}+ \lambda  L_{\mathrm{R}},
\end{align}
where the first term is the negative log-likelihood of assumed Negative Binomial distribution $\mathrm{NB\left(\bar{\mu}^{\mathrm{b}}_g \left(\mathcal{X}^{(n)} \right), \theta^{\mathrm{b}}_g \right)}$ in Equation \ref{eq:dist}, and the second term is the regularization term, which softly takes consistency between single-cell deconvolution results. The $\lambda$ is hyperparameter.
\begin{align}
    L_{\mathrm{NB}} \left(\mathbf{E}^\mathrm{b}, \bar{\boldsymbol{\mu}}^\mathrm{b}, \boldsymbol{\theta}^\mathrm{b} \right)  =& - \frac{1}{N_m} \sum_{n=1}^{N_{\mathrm{m}}} \sum_{g=1}^{N_{\mathrm{ge}}} \log \Pr \left(e^{\mathrm{b} (n)}_g|\mathcal{X}^{(n)} \right),  \notag
\end{align}
where \(\mathbf{E}^\mathrm{b} = \{ \mathbf{e}^{\mathrm{b}(n)} \}^{N_{\mathrm{m}}}_{n=1}\),
\(\bar{\boldsymbol{\mu}}^\mathrm{b} = \left\{\bar{\mu}^\mathrm{b} \left(\mathcal{X}^{(n)} \right) \right\}^{N_\mathrm{m}}_{n=1} \),
\(\boldsymbol{\theta}^\mathrm{b} = [ {\theta}^\mathrm{b}_g ]_{g=1}^{N_{\mathrm{ge}}} \) represent dispersion parameter of Negative Binomial distribution,
$N_{\mathrm{m}}$ denotes the number of the slide in mini-batch.
See Supplementary Material \ref{sec:sup_NB_explanation} for the detailed expression of $L_\mathbf{NB}$

For the regularization term, we take consistency between the cell-type prototype and proportion.
\begin{align}
L_{\mathrm{R}}
(\bar{T}^0, \bar{T}, \mathbf{W}, \bar{\mathbf{W}})
&=
\left\| \bar{T}^0 - \bar{T} \right\|^2 \nonumber \\
&\quad
+ \mathbb{E}_{n}\!\left[
    \left\| \mathbf{W}^{(n)} - \bar{\mathbf{W}}^{(n)} \right\|^2
\right],
\end{align}
where $\mathbf{W}^{(n)}$ denotes the compositional cell-type proportion for the $n$-th sample estimated by deconvolution method \cite{kleshchevnikov2022cell2location}, and $\bar{\mathbf{W}}^{(n)} = \frac{1}{N_{\mathrm{p}}}\sum_i w(\mathbf{x}_i^{(n)})$ represents the mean estimated weight, and $\bar{T}^{0}$ denotes the initial cell-type prototypes.
Our method adjusts the prototype and proportions in a data-driven manner to optimize the reconstruction objective and achieves the estimation of slide-level gene expression with cell-level information.

\subsection{Extension to Patch-Level Estimation}
\label{sec:patch-level}
In spatial transcriptomics (ST), the goal of gene expression estimation is to predict expression levels for each patch image.
Our proposed CPNN framework can be adapted to this task because it aggregates patch-level predictions to infer the slide-level expression.
The main modification lies in the loss function: since the observed ST data tend to be noisy and unstable, we replace the negative binomial loss, which depends on the absolute expression value, $L_{\mathrm{NB}}$ with a Pearson correlation–based loss, which focus on the relative relationships among patches.

For each patch image $\mathbf{x}_i^{(n)}$, the expression is estimated using cell-level prototypes derived following the procedure described in Section \ref{sec:formulation}:
$\hat{e}_{g, i}^{\mathrm{st}(n)} = \alpha_g  \sum_c w \left(\mathbf{x}_i^{(n)} \right)_{c} \bar{T}_{c,g} + \beta_g$, where $\hat{e}_{g, i}^{\mathrm{st}(n)}$ denotes the predicted expression of gene g for the patch. The model parameters are optimized by minimizing the Pearson correlation coefficient (PCC) loss between the predicted and observed expression across all genes with regularization loss for only prototype.
In addition, by leveraging an existing module for estimating $w$, the framework can be seamlessly integrated as a plug-in component within existing analysis pipelines.

\section{Experiments}
We validate the effectiveness of our method in both slide- and patch-level settings.

\begin{table*}[t]
    \caption{Comparison performance of slide-level gene expression estimation on TCGA-BRCA, KIRC, and LUAD. MIL: multiple instance learning, MIR: multiple instance regression, PCC: Pearson correlation coefficient, and SCC: spearman correlation coefficient}
    \label{tab:comparisons}
    \centering
    \begin{tabular}{cc  c c c c c c c c c c} \toprule
         &\multirow{2}{*}{Method} &\multirow{2}{*}{References}& \multicolumn{2}{c}{BRCA} & \multicolumn{2}{c}{KIRC} & \multicolumn{2}{c}{LUAD}  \\ \cmidrule(r){4-5}  \cmidrule(r){6-7} \cmidrule(r){8-9} 
           &  & & PCC & SCC & PCC & SCC & PCC & SCC \\ \midrule
        
        \multirow{8}{*}{\rotatebox{90}{MIL}} &Max \cite{wang2018revisiting}& Pattern Recog. 2018  & 0.219 & 0.225 & 0.162 & 0.161 & 0.170 & 0.173 \\
        &Mean \cite{wang2018revisiting} & Pattern Recog. 2018 & 0.247 & 0.263 & 0.211 & 0.217 & 0.203 & 0.213 \\
        &AbMIL \cite{ilse2018attention} & ICML 2018 & 0.280 & 0.310 & 0.258 & 0.260 & 0.261 & 0.270 \\
        &ILRA \cite{xiang2023exploring} & ICLR 2023 & 0.291 & 0.310 & 0.261 & 0.276 & 0.245 & 0.283 \\
        &S4MIL \cite{fillioux2023structured} & MICCAI2023& 0.292 & 0.301 & 0.225 & 0.231 & 0.243 & 0.252 \\
        &MambaMIL \cite{yang2024mambamil} & MICCAI 2024 & 0.293 & 0.309 & 0.212 & 0.224 & \bf{0.280} & 0.284 \\
        &SRMambaMIL \cite{yang2024mambamil} & MICCAI 2024 & 0.304 & 0.311 & 0.244 & 0.251 & 0.276 & 0.286 \\
        &2DMamba \cite{zhang20252dmamba} & CVPR2025 & 0.276 & 0.286 & 0.219 & 0.229 & 0.249 & 0.251 \\

        \hdashline 
        
        \multirow{6}{*}{\rotatebox{90}{MIR}}&HE2RNA \cite{schmauch2020deep} & Nat. Commun. 2020 & 0.280 & 0.298 & 0.274 & 0.292 & 0.267 & 0.285 \\
        &AbReg \cite{graziani2022attention} & MICCAIW 2022 & 0.267 & 0.273 & 0.242 & 0.244 & 0.243 & 0.251 \\
        &tRNAformer \cite{Alsaafin2022LearningTP}& Commun. Biol. 2022 & 0.291 & 0.302 & 0.283 & 0.291 & 0.262 & 0.282 \\
        &MOSBY \cite{csenbabaouglu2024mosby}& Sci. Rep. 2024 & 0.286 & 0.314 & 0.241 & 0.271 & 0.259 & 0.274 \\

        &SEQUOIA VIS \cite{pizurica2024digital} & Nat. Commun. 2024 & 0.285 & 0.295 & 0.267 & 0.272 & 0.252 & 0.258 \\

        &Ours & - &\bf{0.304} & \bf{0.338} & \bf{0.291} & \bf{0.318} & 0.271 & \bf{0.304} \\

        \bottomrule
        \hline
    \end{tabular}
\end{table*}

\noindent
{\bf Evaluation Metrics.}
Since the scaling bias of the sequencing technique can cause fluctuations in the absolute gene expression values, we used the Pearson and Spearman correlation coefficients (PCC, SCC) for the evaluation metrics. This metric was chosen to consider practical applications like Differentially Expressed Genes analysis \cite{love2014moderated}, which analyzes the relative expression difference between samples as the downstream task.

\subsection{Evaluation on Slide-Level Estimation}
\noindent
{\bf Experimental Setup.}
We used CLAM implementation \cite{lu2021data} for MIL-based preprocessing to make a patch and CONCH \cite{lu2024visual} for the image encoder $f_{\mathrm{f}}$. For the $f_{\mathrm{mlp}}$, we adopt the projector architecture of LLaVA 1.5 \cite{liu2024improved}.
For the Negative Binomial Regression and deconvolution, we used the implementation of cell2location \cite{kleshchevnikov2022cell2location} with default parameters.
We trained our method using the AdamW optimizer \cite{loshchilov2018decoupled} with a mini-batch size of 16 for 500 epochs. Early stopping was applied based on validation loss, using a validation set containing 30 patients. The learning rate was set to $1\times10^{-3}$ and the regularization coefficient $\lambda$ to $1\times10^{3}$. For each dataset, we performed 4-fold cross-validation

\noindent
{\bf Datasets:}
To evaluate the effectiveness of the proposed method, we validated our approach using three publicly available datasets from TCGA: BRCA, KIRC, and LUAD.
Details of preprocessing for both slide-level and cell-level datasets are provided in Supplementary Material \ref{sec:sup_dataset}.
An overview of each dataset is summarized below. {\bf BRCA:}
We collected 1,474 paired samples of whole-slide images (WSIs) and bulk gene expression profiles from TCGA-BRCA.
For the single-cell dataset, 77,839 cells were obtained from \cite{wu2021single}.
Cell-type annotations were assigned to clusters based on known marker gene expression, resulting in 49 cell types.
A total of 14,047 genes shared between the bulk and single-cell datasets were used for gene expression estimation.
{\bf KIRC:} 
We collected 681 paired samples from TCGA-KIRC.
A total of 265,825 cells were obtained from the RCC single-cell dataset \cite{li2022mapping}.
Cell-type labels (106 types) were assigned following the same procedure as BRCA.
We selected 14,300 genes common to both datasets for downstream estimation.
{\bf LUAD:}
We collected 756 paired samples from TCGA-LUAD.
For the single-cell data, 171,128 cells were obtained from \cite{kim2020single}.
In total, 49 cell types were annotated.
A set of 14,520 genes shared by the bulk and single-cell datasets was used for estimation.

\begin{table*}[t]
    \centering
    \caption{Comparison of methods across CSCC, Her2st, and STNet datasets. Each (\textcolor{green!60!black}{+digit} or \textcolor{red}{-digit}) shows the difference from the baseline}\label{tab:stcomparison}  
    \scalebox{0.95}{
    \begin{tabular}{lc cc cc cc}
    \toprule
     \multirow{2}{*}{Method} & \multirow{2}{*}{Ref.} & \multicolumn{2}{c}{CSCC \cite{ji2020multimodal}} & \multicolumn{2}{c}{Her2st \cite{andersson2020spatial}} & \multicolumn{2}{c}{STNet \cite{he2020integrating}} \\
    \cmidrule(r){3-4} \cmidrule(r){5-6}  \cmidrule(r){7-8} 
     & & PCC & SCC & PCC & SCC & PCC & SCC \\
    \midrule
    BLEEP \cite{xie2023spatially} & NeurIPS 2023& -0.0050 & -0.0024 & 0.0212 & 0.0321 & -0.0051	& -0.0062  \\
    EGN \cite{yang2023exemplar}& WACV 2023 & 0.0552 & 0.1340 &  0.0941&  0.0479 & 0.0570 & 0.0272 \\ \hdashline
    STNet \cite{he2020integrating} & Nat. Biomed.  & 0.0856 & 0.1156 & 0.0849 & 0.0772 & 0.0514 & 0.0536 \\
    w/ Ours & Eng. 2020 & 
    0.1008 {\tiny \textcolor{green!60!black}{(+0.0152)}} & 
    0.1662 {\tiny \textcolor{green!60!black}{(+0.0506)}} & 
    0.0825 {\tiny \textcolor{red!70!black}{(-0.0024)}} & 
    0.1111 {\tiny \textcolor{green!60!black}{(+0.0339)}} & 
    0.0532 {\tiny \textcolor{green!60!black}{(+0.0018)}} & 
    \textbf{0.0623} {\tiny \textcolor{green!60!black}{(+0.0087)}} \\ \hdashline
    TRIPLEX \cite{chung2024accurate} & \multirow{2}{*}{CVPR 2024} & 
    0.0895 & 0.1239 & \textbf{0.0900} & 0.0861 & \textbf{0.0606} & 0.0546 \\
    w/ Ours & & 
    \textbf{0.1116} {\tiny \textcolor{green!60!black}{(+0.0221)}} & 
    \textbf{0.1821} {\tiny \textcolor{green!60!black}{(+0.0582)}} & 
    0.0865 {\tiny \textcolor{red!70!black}{(-0.0035)}} & 
    \textbf{0.1194} {\tiny \textcolor{green!60!black}{(+0.0333)}} & 
    0.0523 {\tiny \textcolor{red!70!black}{(-0.0083)}} & 
    0.0621 {\tiny \textcolor{green!60!black}{(+0.0075)}} \\
    \bottomrule
    \end{tabular}
}
\end{table*}

\noindent
{\bf Comparisons on Slide-Level Estimation.}
To evaluate the effectiveness of the proposed method, we compared it with thirteen existing approaches, including multiple-instance learning (MIL) and multiple-instance regression (MIR) methods.
MIL baselines include pooling-based models (\textbf{Max}, \textbf{Mean} \cite{wang2018revisiting}), attention-based models (\textbf{AbMIL}, \textbf{ILRA}), and recent state-space models (\textbf{S4MIL} \cite{fillioux2023structured}, \textbf{MambaMIL}, \textbf{SRMambaMIL} \cite{yang2024mambamil}, \textbf{2DMamba} \cite{zhang20252dmamba}).
For MIR, we included transformer- and clustering-based methods such as \textbf{HE2RNA} \cite{schmauch2020deep}, \textbf{AbReg} \cite{graziani2022attention}, \textbf{tRNAformer} \cite{Alsaafin2022LearningTP}, \textbf{MOSBY} \cite{csenbabaouglu2024mosby}, and \textbf{SEQUOIA VIS} \cite{pizurica2024digital}.
All models used the same feature extractor (CONCH) and identical training settings for fair comparison.

Table~\ref{tab:comparisons} summarizes the overall results. 
Our proposed method achieves state-of-the-art (SOTA) performance on the SCC metric across all datasets. 
For the PCC metric, it attains the best performance on BRCA and KIRC and ranks third on LUAD.

Among MIL-based approaches, transformer- and state-space–based models such as ILRA, S4MIL, and MambaMIL outperform conventional pooling methods (Max, Mean), indicating that attention-like mechanisms capture long-range contextual dependencies more effectively. 
However, even with these advanced aggregators, the performance gains remain limited.

For MIR-based methods, models specifically designed for regression—such as HE2RNA and tRNAformer—achieve slightly higher correlations than MIL-based approaches, particularly on the KIRC and LUAD datasets. 
These models benefit from instance selection and transformer-based feature learning. 
Nevertheless, estimating high-dimensional gene expression remains challenging without appropriate regularization.

Our proposed method achieved the highest overall performance across all datasets. 
These results demonstrate that our prototype-guided framework effectively supports high-dimensional gene expression estimation. 
Because each pre-defined prototype provide stable gene–gene covariation patterns, it serves as an implicit regularizer within the prediction space. 
Importantly, the ability to leverage cell-level information, even in the presence of a modality gap between single-cell and histology data, demonstrates the robustness and versatility of our framework.

\subsection{Evaluation on Patch-Level Estimation}
\noindent
{\bf Experimental Setup and Datasets.}
Data preprocessing followed the protocol described in \cite{chung2024accurate}. 
We used three spatial transcriptomics datasets: a cutaneous squamous cell carcinoma (CSCC) dataset \cite{ji2020multimodal} and two breast cancer datasets, Her2st \cite{andersson2020spatial} and STNet \cite{he2020integrating}. 
The CSCC dataset includes both single-cell and spatial transcriptomics modalities; the corresponding single-cell data were obtained from GSE144240 \cite{ji2020multimodal}. 
For Her2st and STNet, we used the GSE176078 dataset \cite{wu2021single}, which was also used in the BRCA experiments. 
Unlike prior implementations, we excluded spatial smoothing, as it may introduce bias in downstream analyses \cite{ganguly2025merge}. 
Target genes were selected using the same procedure used for slide-level estimation: after filtering both single-cell and ST data, only genes shared across the two modalities were retained. 
This resulted in 10,118 genes for CSCC, 9,279 for Her2st, and 7,716 for STNet.
For each dataset, model training was performed using patient-level leave-one-out cross-validation to ensure robust subject-level evaluation.

\noindent
\textbf{Comparisons on Patch-Level Estimation.}
We compared our method with four representative benchmarks: 
\textbf{BLEEP} \cite{xie2023spatially}, which employs multimodal contrastive learning for gene expression prediction; 
\textbf{EGN} \cite{yang2023exemplar}, which introduces an exemplar-guided estimation strategy; 
\textbf{STNet} \cite{he2020integrating}, which fine-tunes DenseNet-121 with a linear regression head; 
and \textbf{TRIPLEX} \cite{chung2024accurate}, which uses a multi-scale feature aggregation framework.
Our method can be integrated into feature-extraction–based models by replacing their feature extractor $f_{\mathrm{f}}$. 
To demonstrate this flexibility, we incorporated CPNN into STNet and TRIPLEX and evaluated their performance using the same training protocol.

Table~\ref{tab:stcomparison} summarizes the results. 
Integrating CPNN consistently improved SCC performance across all datasets, with the largest gain observed on the CSCC dataset. 
Because the CSCC dataset contains matched single-cell and spatial transcriptomics data from the same patients, modality alignment is preserved, allowing the prototype information to be utilized more effectively.

Despite the inherent differences in expression scales across modalities, the proposed method maintained comparable PCC while substantially improving SCC. 
These results indicate that incorporating cell-type prototypes provides meaningful regularization for spatial gene expression prediction, even under multimodal settings, and can enhance the performance of existing patch-level estimation frameworks.

\begin{table}[t]
    \centering
    \caption{
        Ablation study results.
        PI: prototype initialization of $\bar{T}$, 
        MC: modality correction using $\alpha$ and $\beta$, 
        U: prototype update of $\bar{T}$, 
        R: regularization loss.
    }
    \label{tab:ablation}
    \begin{tabular}{lccccc}
        \toprule
        \textbf{Settings} & \textbf{PI} & \textbf{MC} & \textbf{U} & \textbf{R} & \textbf{SCC} \\
        \midrule
        w/o PI, MC, R & \ding{55} & \ding{55} & \checkmark & \ding{55} & 0.305 \\
        w/o MC, U, R  & \checkmark & \ding{55} & \ding{55} & \ding{55} & 0.174 \\
        w/o U, R      & \checkmark & \checkmark & \ding{55} & \ding{55} & 0.248 \\
        w/o R         & \checkmark & \checkmark & \checkmark & \ding{55} &  0.336\\
        \textbf{Ours} & \checkmark & \checkmark & \checkmark & \checkmark & \textbf{0.338} \\
        \bottomrule
    \end{tabular}
\end{table}
\subsection{Ablation Study}
\noindent
{\bf Effect of Individual Modules.} We assess the module-wise effectiveness of the proposed framework in the slide-level gene expression estimation setting using the BRCA dataset ({\it i.e.,} an ablation over each model component). Table \ref{tab:ablation} shows the results of validating each module. PI: prototype-informed, which uses estimated $\bar{T}$ for initial value of prototype, MC: modality correction by $\alpha$ and $\beta$, U: whether to update prototype $T$, and R: using regularization loss. "w/o PI, MC, R" refers to the case where proportion and prototype were trained from scratch with $L_{\mathbf{NB}}$. 

The performance of "w/o PI, MC, R" is similar to MOSBY, which takes a summation of the estimated gene expression for each patch. Our findings suggest that without any guidance of prototype and regularization, it shows similar performance to MOSBY and yields no performance gain. The result demonstrates that our prototype-informed approach contributes to improving performance. 
There are biases between single-cell and bulk data due to multiple factors. Consequently, using a prototype derived from single-cell data without updating it results in performance degradation relative to scratch (w/o MC, U, R).
To address the bias between the two data datasets, we introduced modality correction using factors $\vec{\alpha}$ and $\vec{\beta}$, leading to a performance gain (w/o U, R). 
However, the modality correction alone was insufficient to bridge the modality gap, limiting overall improvement fully.
Updating the prototype ("w/o R") improved performance compared to training from scratch, indicating that initializing with our cell-type prototype-informed model enhances generalization. 
While the regularization term does not lead to significant performance gains, it enforces consistency with the deconvolution results, which in turn preserves the interpretability of the estimated cell-type compositions.

\begin{table}[t]
    \centering
    \caption{Effect of cell-type label granularity on model performance (SCC).}
    \label{tab:granularity}
    \begin{tabular}{cccccc} 
          \toprule
           Coarse (8)& medium (29) & fine (49) \\ \midrule
           0.317 & 0.336& 0.338\\
          \bottomrule
    \end{tabular}
\end{table}

\noindent
\textbf{Effect of Cell-Type Label Granularity.}
Cell-type annotations vary in granularity. For instance, T cells can be subdivided into finer categories such as CD4\textsuperscript{+} T cells and conventional T cells. Single-cell datasets often provide labels at multiple resolutions, and the chosen level of granularity may influence model performance. Coarse labels may obscure biologically meaningful heterogeneity, whereas overly fine-grained labels may produce categories that are not visually distinguishable in histology, resulting in limited performance gains.

To assess the effect of label granularity, we constructed prototypes using BRCA single-cell data under three labeling schemes—coarse, medium, and fine—containing 9, 29, and 49 cell types, respectively. 
Table~\ref{tab:granularity} summarizes the results in terms of SCC.
Overall, the model remained robust across medium and fine levels of granularity, with only minor variations in SCC and PCC. 
Performance degradation was observed primarily with coarse labels, likely due to the reduced number of prototypes limiting the model's representational capacity. 
These results suggest that moderate granularity provides a favorable balance between biological resolution and model flexibility.

\noindent
{\bf Biological Validation of Compositional Weights.}
By integrating cell-type prototypes into the model, we obtain not only predicted gene expression values but also the corresponding prototype weights. These weights provide cell-type–level explanations for the predictions, thereby enhancing interpretability.
To evaluate whether the estimated compositional weights are biologically meaningful, we analyzed the BRCA dataset and examined the prototype weights predicted for each whole-slide image. We focused on cancer-related prototypes derived from single-cell data and visualized their average weights across slides.

\begin{figure}[t]
    \centering
    \includegraphics[width=0.93\linewidth]{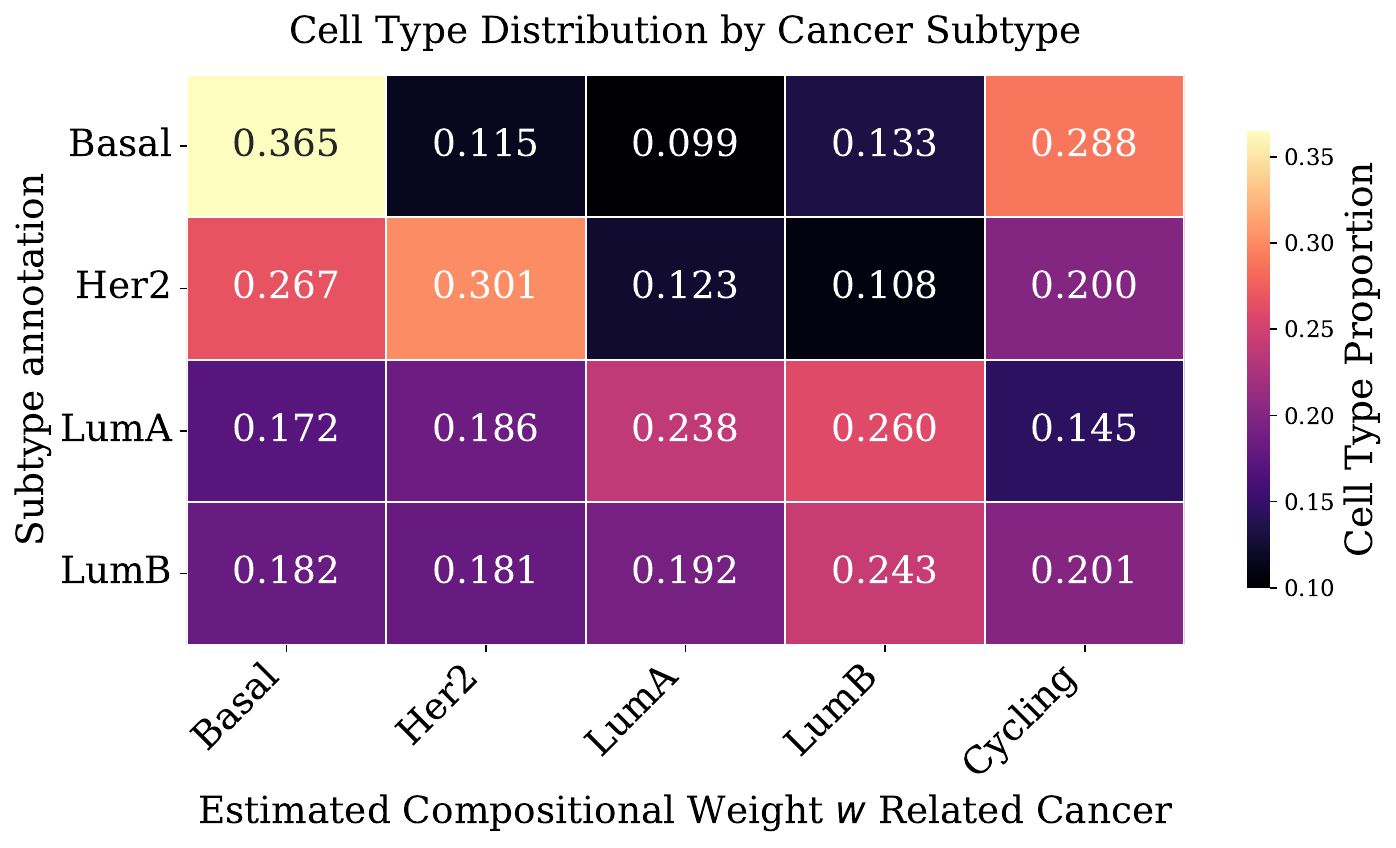}
    \caption{Average weight for each subtype of BRCA slides. We visualized the estimated weights for the cancer-related prototypes.}
    \label{fig:proportion}
\end{figure}

Figure~\ref{fig:proportion} shows the average weights predicted by CPNN for each BRCA subtype. In addition to the subtype-associated prototypes, we also visualize the Cycling prototype, which represents proliferative tumor programs and is informative for distinguishing tumor aggressiveness.

First, focusing on the subtype-specific prototypes (excluding Cycling), we find that each subtype exhibits a high probability for the compositional weight corresponding to its expected cell-type program. Despite the absence of subtype supervision during training, the model assigns higher weights to prototypes that align with known BRCA subtype characteristics. In LumA tumors, both LumA and LumB prototypes show similar activation levels. This observation is consistent with previous studies \cite{wu2021single} and likely reflects the similar expression profiles shared by these two subtypes.
The Cycling prototype receives the highest weights in Basal-like tumors, consistent with their highly proliferative phenotype. Cycling weights are also higher in LumB tumors than in LumA tumors, reflecting the greater proliferative activity known for LumB.

Collectively, these results demonstrate that the inferred compositional weights reveal which cell types drive the predicted expression patterns, thereby providing clear and biologically grounded interpretability.

\section{Conclusion}

In this paper, we proposed a Cell-type Prototype-informed Neural Network (CPNN) for gene expression estimation from whole-slide images. 
To leverage cell-level datasets despite their modality gap and severe noise, we introduced cell-type prototypes, which represent stable gene expression profiles for each cell type. 
Across three slide-level and three patch-level datasets, CPNN consistently outperformed existing methods, demonstrating its robustness and general applicability. 
Moreover, the prototype decomposition provides interpretable cell-type contributions, offering biologically meaningful insights that cannot be obtained through conventional feature-based approaches. 
These results suggest that incorporating cell-level information through cell-type prototypes is a promising direction for improving both accuracy and interpretability in computational gene expression prediction.

\section*{Acknowledgment}
Funding in direct support of this work: JSPS KAKEN JP24KJ2205, 23H03862, and JP23K16991, JST ACT-X JPMJAX23CR, JST ASPIRE JPMJAP2403, AMED grant JP24ama221609h0001(P-PROMOTE) (to YK) and JP25gm7010014h0001(PRIME), National Cancer Center Research and Development Fund 2024-A-6 (to YK).
We used ABCI 3.0 provided by AIST and AIST Solutions.

{
    \small
    \bibliographystyle{ieeenat_fullname}
    \bibliography{main}

@String(CVPR= {IEEE Conf. Comput. Vis. Pattern Recog.})

@String(ICLR = {Int. Conf. Learn. Represent.})

@String(CVPR  = {CVPR})

@String(ICLR  = {ICLR})

@article{wu2021single,
  title={{A single-cell and spatially resolved atlas of human breast cancers}},
  author={Wu, Sunny Z and Al-Eryani, Ghamdan and Roden, Daniel Lee and Junankar, Simon and Harvey, Kate and Andersson, Alma and Thennavan, Aatish and Wang, Chenfei and Torpy, James R and Bartonicek, Nenad and others},
  journal={Nature genetics},
  volume={53},
  number={9},
  pages={1334--1347},
  year={2021},
  publisher={Nature Publishing Group US New York}
}

@article{zheng2024digital,
  title={{Digital profiling of cancer transcriptomes from histology images with grouped vision attention}},
  author={Zheng, Yuanning and Pizurica, Marija and Carrillo-Perez, Francisco and Noor, Humaira and Yao, Wei and Wohlfart, Christian and Marchal, Kathleen and Vladimirova, Antoaneta and Gevaert, Olivier},
  journal={bioRxiv},
  pages={2023--09},
  year={2024}
}

@article{pizurica2024digital,
  title={{Digital profiling of gene expression from histology images with linearized attention}},
  author={Pizurica, Marija and Zheng, Yuanning and Carrillo-Perez, Francisco and Noor, Humaira and Yao, Wei and Wohlfart, Christian and Vladimirova, Antoaneta and Marchal, Kathleen and Gevaert, Olivier},
  journal={Nat. Commun.},
  volume={15},
  number={1},
  pages={9886},
  year={2024},
  publisher={Nature Publishing Group UK London}
}

@article{csenbabaouglu2024mosby,
  title={{MOSBY enables multi-omic inference and spatial biomarker discovery from whole slide images}},
  author={{\c{S}}enbabao{\u{g}}lu, Yasin and Prabhakar, Vignesh and Khormali, Aminollah and Eastham, Jeff and Liu, Evan and Warner, Elisa and Nabet, Barzin and Srivastava, Minu and Ballinger, Marcus and Liu, Kai},
  journal={Sci. Rep.},
  volume={14},
  number={1},
  pages={18271},
  year={2024},
  publisher={Nature Publishing Group UK London}
}

@inproceedings{graziani2022attention,
  title={{Attention-based interpretable regression of gene expression in histology}},
  author={Graziani, Mara and Marini, Niccol{\`o} and Deutschmann, Nicolas and Janakarajan, Nikita and M{\"u}ller, Henning and Mart{\'\i}nez, Mar{\'\i}a Rodr{\'\i}guez},
  booktitle={MICCAIW},
  pages={44--60},
  year={2022},
  organization={Springer}
}

@article{schmauch2020deep,
  title={A deep learning model to predict RNA-Seq expression of tumours from whole slide images},
  author={Schmauch, Beno{\^\i}t and Romagnoni, Alberto and Pronier, Elodie and Saillard, Charlie and Maill{\'e}, Pascale and Calderaro, Julien and Kamoun, Aur{\'e}lie and Sefta, Meriem and Toldo, Sylvain and Zaslavskiy, Mikhail and others},
  journal={Nat. Commun.},
  volume={11},
  number={1},
  pages={3877},
  year={2020},
  publisher={Nature Publishing Group UK London}
}

@article{Alsaafin2022LearningTP,
  title={{Learning to predict RNA sequence expressions from whole slide images with applications for search and classification}},
  author={Areej Alsaafin and Amir Safarpoor and Milad Sikaroudi and Jason D. Hipp and Hamid R. Tizhoosh},
  journal={Commun. Biol.},
  year={2022},
  volume={6},
  url={https://api.semanticscholar.org/CorpusID:247762378}
}

@inproceedings{ilse2018attention,
  title={{Attention-based deep multiple instance learning}},
  author={Ilse, Maximilian and Tomczak, Jakub and Welling, Max},
  booktitle={ICML},
  pages={2127--2136},
  year={2018},
  organization={PMLR}
}

@article{wang2018revisiting,
  title={{Revisiting multiple instance neural networks}},
  author={Wang, Xinggang and Yan, Yongluan and Tang, Peng and Bai, Xiang and Liu, Wenyu},
  journal={Nat. Med.Pattern Recognit.},
  volume={74},
  pages={15--24},
  year={2018},
  publisher={Elsevier}
}

@inproceedings{xiang2023exploring,
  title={{Exploring low-rank property in multiple instance learning for whole slide image classification}},
  author={Xiang, Jinxi and Zhang, Jun},
  booktitle={ICLR},
  year={2023}
}

@article{lu2021data,
  title={Data-efficient and weakly supervised computational pathology on whole-slide images},
  author={Lu, Ming Y and Williamson, Drew FK and Chen, Tiffany Y and Chen, Richard J and Barbieri, Matteo and Mahmood, Faisal},
  journal={Nat. Biomed. Eng.},
  volume={5},
  number={6},
  pages={555--570},
  year={2021},
  publisher={Nature Publishing Group}
}

@article{kleshchevnikov2022cell2location,
  title={{Cell2location maps fine-grained cell types in spatial transcriptomics}},
  author={Kleshchevnikov, Vitalii and Shmatko, Artem and Dann, Emma and Aivazidis, Alexander and King, Hamish W and Li, Tong and Elmentaite, Rasa and Lomakin, Artem and Kedlian, Veronika and Gayoso, Adam and others},
  journal={Nat. Biotechnol.},
  volume={40},
  number={5},
  pages={661--671},
  year={2022},
  publisher={Nature Publishing Group US New York}
}

@article{mortazavi2008mapping,
  title={{Mapping and quantifying mammalian transcriptomes by RNA-Seq}},
  author={Mortazavi, Ali and Williams, Brian A and McCue, Kenneth and Schaeffer, Lorian and Wold, Barbara},
  journal={Nat. methods},
  volume={5},
  number={7},
  pages={621--628},
  year={2008},
  publisher={Nature Publishing Group}
}

@article{marx2021method,
  title={{Method of the Year: spatially resolved transcriptomics}},
  author={Marx, Vivien},
  journal={Nat. Methods},
  volume={18},
  number={1},
  pages={9--14},
  year={2021},
  publisher={Nature Publishing Group US New York}
}

@article{jovic2022single,
  title={{Single-cell RNA sequencing technologies and applications: A brief overview}},
  author={Jovic, Dragomirka and Liang, Xue and Zeng, Hua and Lin, Lin and Xu, Fengping and Luo, Yonglun},
  journal={Clin. Transl. Med.},
  volume={12},
  number={3},
  pages={e694},
  year={2022},
  publisher={Wiley Online Library}
}

@article{love2014moderated,
  title={{Moderated estimation of fold change and dispersion for RNA-seq data with DESeq2}},
  author={Love, Michael I and Huber, Wolfgang and Anders, Simon},
  journal={Genome biology},
  volume={15},
  pages={1--21},
  year={2014},
  publisher={Springer}
}

@article{wang2019bulk,
  title={Bulk tissue cell type deconvolution with multi-subject single-cell expression reference},
  author={Wang, Xuran and Park, Jihwan and Susztak, Katalin and Zhang, Nancy R and Li, Mingyao},
  journal={Nat. Commun.},
  volume={10},
  number={1},
  pages={380},
  year={2019},
  publisher={Nature Publishing Group UK London}
}

@article{gynter2023deconv,
  title={{DeconV: Probabilistic Cell Type Deconvolution from Bulk RNA-sequencing Data}},
  author={Gynter, Artur and Meistermann, Dimitri and L{\"a}hdesm{\"a}ki, Harri and Kilpinen, Helena},
  journal={bioRxiv},
  pages={2023--12},
  year={2023},
  publisher={Cold Spring Harbor Laboratory}
}

@article{chu2022cell,
  title={{Cell type and gene expression deconvolution with BayesPrism enables Bayesian integrative analysis across bulk and single-cell RNA sequencing in oncology}},
  author={Chu, Tinyi and Wang, Zhong and Pe’er, Dana and Danko, Charles G},
  journal={Nat. Cancer},
  volume={3},
  number={4},
  pages={505--517},
  year={2022},
  publisher={Nature Publishing Group US New York}
}

@article{pang2021leveraging,
  title={Leveraging information in spatial transcriptomics to predict super-resolution gene expression from histology images in tumors},
  author={Pang, Minxing and Su, Kenong and Li, Mingyao},
  journal={BioRxiv},
  pages={2021--11},
  year={2021},
  publisher={Cold Spring Harbor Laboratory}
}

@article{li2022mapping,
  title={Mapping single-cell transcriptomes in the intra-tumoral and associated territories of kidney cancer},
  author={Li, Ruoyan and Ferdinand, John R and Loudon, Kevin W and Bowyer, Georgina S and Laidlaw, Sean and Muyas, Francesc and Mamanova, Lira and Neves, Joana B and Bolt, Liam and Fasouli, Eirini S and others},
  journal={Cancer cell},
  volume={40},
  number={12},
  pages={1583--1599},
  year={2022},
  publisher={Elsevier}
}

@inproceedings{
loshchilov2018decoupled,
title={Decoupled Weight Decay Regularization},
author={Ilya Loshchilov and Frank Hutter},
booktitle={ICLR},
year={2019},
url={https://openreview.net/forum?id=Bkg6RiCqY7},
}

@article{kim2020single,
  title={{Single-cell RNA sequencing demonstrates the molecular and cellular reprogramming of metastatic lung adenocarcinoma}},
  author={Kim, Nayoung and Kim, Hong Kwan and Lee, Kyungjong and Hong, Yourae and Cho, Jong Ho and Choi, Jung Won and Lee, Jung-Il and Suh, Yeon-Lim and Ku, Bo Mi and Eum, Hye Hyeon and others},
  journal={Nat. Commun.},
  volume={11},
  number={1},
  pages={2285},
  year={2020},
  publisher={Nature Publishing Group UK London}
}

@inproceedings{liu2024improved,
  title={Improved baselines with visual instruction tuning},
  author={Liu, Haotian and Li, Chunyuan and Li, Yuheng and Lee, Yong Jae},
  booktitle={CVPR},
  pages={26296--26306},
  year={2024}
}

@article{icgc2020pan,
  author = {ICGC/TCGA Pan-Cancer Analysis of Whole Genomes Consortium},
  title = {Pan-cancer analysis of whole genomes},
  journal = {Nature},
  volume = {578},
  number = {7793},
  pages = {82--93},
  year = {2020},
  publisher = {Nature Publishing Group UK London}
}

@article{barrett2012ncbi,
  title={{NCBI GEO: archive for functional genomics data sets—update}},
  author={Barrett, Tanya and Wilhite, Stephen E and Ledoux, Pierre and Evangelista, Carlos and Kim, Irene F and Tomashevsky, Maxim and Marshall, Kimberly A and Phillippy, Katherine H and Sherman, Patti M and Holko, Michelle and others},
  journal={Nucleic Acids Res.},
  volume={41},
  number={D1},
  pages={D991--D995},
  year={2012},
  publisher={Oxford University Press}
}

@article{Tarhan2023SingleCP,
  title={{Single Cell Portal: an interactive home for single-cell genomics data}},
  author={Leyla Tarhan and Jonathan Bistline and Jean Chang and Bryan Galloway and Emily Hanna and Eric M. Weitz},
  journal={bioRxiv},
  year={2023},
  url={https://api.semanticscholar.org/CorpusID:259976065}
}

@article{Abugessaisa2017SCPortalenHA,
  title={{SCPortalen: human and mouse single-cell centric database}},
  author={Imad Eldin Ali Abugessaisa and Shuhei Noguchi and Michael B{\"o}ttcher and Akira Hasegawa and Tsukasa Kouno and Sachi Kato and Yuhki Tada and Hiroki Ura and Kuniya Abe and Jay W. Shin and Charles Plessy and Piero Carninci and Takeya Kasukawa},
  journal={Nucleic Acids Res.},
  year={2017},
  volume={46},
  pages={D781 - D787},
  url={https://api.semanticscholar.org/CorpusID:24216}
}

@article{he2020integrating,
  title={Integrating spatial gene expression and breast tumour morphology via deep learning},
  author={He, Bryan and Bergenstr{\aa}hle, Ludvig and Stenbeck, Linnea and Abid, Abubakar and Andersson, Alma and Borg, {\AA}ke and Maaskola, Jonas and Lundeberg, Joakim and Zou, James},
  journal={Nat. Biomed. Eng.},
  volume={4},
  number={8},
  pages={827--834},
  year={2020},
  publisher={Nature Publishing Group UK London}
}

@article{zeng2022spatial,
  title={Spatial transcriptomics prediction from histology jointly through transformer and graph neural networks},
  author={Zeng, Yuansong and Wei, Zhuoyi and Yu, Weijiang and Yin, Rui and Yuan, Yuchen and Li, Bingling and Tang, Zhonghui and Lu, Yutong and Yang, Yuedong},
  journal={Brief. Bioinform.},
  volume={23},
  number={5},
  pages={bbac297},
  year={2022},
  publisher={Oxford University Press}
}

@inproceedings{yang2023exemplar,
  title={Exemplar guided deep neural network for spatial transcriptomics analysis of gene expression prediction},
  author={Yang, Yan and Hossain, Md Zakir and Stone, Eric A and Rahman, Shafin},
  booktitle={WACV},
  pages={5039--5048},
  year={2023}
}

@inproceedings{chung2024accurate,
  title={Accurate spatial gene expression prediction by integrating multi-resolution features},
  author={Chung, Youngmin and Ha, Ji Hun and Im, Kyeong Chan and Lee, Joo Sang},
  booktitle=CVPR,
  pages={11591--11600},
  year={2024}
}

@article{lopez2018deep,
  title={Deep generative modeling for single-cell transcriptomics},
  author={Lopez, Romain and Regier, Jeffrey and Cole, Michael B and Jordan, Michael I and Yosef, Nir},
  journal={Nat. Methods},
  volume={15},
  number={12},
  pages={1053--1058},
  year={2018},
  publisher={Nature Publishing Group US New York}
}

@article{toscano2024pinns,
  title={From pinns to pikans: Recent advances in physics-informed machine learning},
  author={Toscano, Juan Diego and Oommen, Vivek and Varghese, Alan John and Zou, Zongren and Daryakenari, Nazanin Ahmadi and Wu, Chenxi and Karniadakis, George Em},
  journal={arXiv preprint arXiv:2410.13228},
  year={2024}
}

@article{kamali2023elasticity,
  title={Elasticity imaging using physics-informed neural networks: Spatial discovery of elastic modulus and Poisson's ratio},
  author={Kamali, Ali and Sarabian, Mohammad and Laksari, Kaveh},
  journal={Acta Biomater.},
  volume={155},
  pages={400--409},
  year={2023},
  publisher={Elsevier}
}

@article{kojima2024single,
  title={Single-cell colocalization analysis using a deep generative model},
  author={Kojima, Yasuhiro and Mii, Shinji and Hayashi, Shuto and Hirose, Haruka and Ishikawa, Masato and Akiyama, Masashi and Enomoto, Atsushi and Shimamura, Teppei},
  journal={Cell Systems},
  volume={15},
  number={2},
  pages={180--192},
  year={2024},
  publisher={Elsevier}
}

@article{jo2024density,
  title={Density physics-informed neural networks reveal sources of cell heterogeneity in signal transduction},
  author={Jo, Hyeontae and Hong, Hyukpyo and Hwang, Hyung Ju and Chang, Won and Kim, Jae Kyoung},
  journal={Patterns},
  volume={5},
  number={2},
  year={2024},
  publisher={Elsevier}
}

@article{yasuda2022super,
  title={Super-resolution of near-surface temperature utilizing physical quantities for real-time prediction of urban micrometeorology},
  author={Yasuda, Yuki and Onishi, Ryo and Hirokawa, Yuichi and Kolomenskiy, Dmitry and Sugiyama, Daisuke},
  journal={Building and Environment},
  volume={209},
  pages={108597},
  year={2022},
  publisher={Elsevier}
}

@inproceedings{yang2024mambamil,
  title={{MambaMIL}: Enhancing long sequence modeling with sequence reordering in computational pathology},
  author={Yang, Shu and Wang, Yihui and Chen, Hao},
  booktitle={MICCAI},
  pages={296--306},
  year={2024},
  organization={Springer}
}

@inproceedings{fillioux2023structured,
  title={{Structured state space models for multiple instance learning in digital pathology}},
  author={Fillioux, Leo and Boyd, Joseph and Vakalopoulou, Maria and Courn{\`e}de, Paul-Henry and Christodoulidis, Stergios},
  booktitle={MICCAI},
  pages={594--604},
  year={2023},
  organization={Springer}
}

@inproceedings{zhang20252dmamba,
  title={{2dMamba: Efficient state space model for image representation with applications on giga-pixel whole slide image classification}},
  author={Zhang, Jingwei and Nguyen, Anh Tien and Han, Xi and Trinh, Vincent Quoc-Huy and Qin, Hong and Samaras, Dimitris and Hosseini, Mahdi S},
  booktitle={CVPR},
  pages={3583--3592},
  year={2025}
}

@article{lu2024visual,
  title={{A visual-language foundation model for computational pathology}},
  author={Lu, Ming Y and Chen, Bowen and Williamson, Drew FK and Chen, Richard J and Liang, Ivy and Ding, Tong and Jaume, Guillaume and Odintsov, Igor and Le, Long Phi and Gerber, Georg and others},
  journal={Nat. Med.},
  volume={30},
  number={3},
  pages={863--874},
  year={2024},
  publisher={Nature Publishing Group US New York}
}

@article{xie2023spatially,
  title={{Spatially resolved gene expression prediction from histology images via bi-modal contrastive learning}},
  author={Xie, Ronald and Pang, Kuan and Chung, Sai and Perciani, Catia and MacParland, Sonya and Wang, Bo and Bader, Gary},
  journal={NeurIPS},
  volume={36},
  pages={70626--70637},
  year={2023}
}

@article{ji2020multimodal,
  title={{Multimodal analysis of composition and spatial architecture in human squamous cell carcinoma}},
  author={Ji, Andrew L and Rubin, Adam J and Thrane, Kim and Jiang, Sizun and Reynolds, David L and Meyers, Robin M and Guo, Margaret G and George, Benson M and Mollbrink, Annelie and Bergenstr{\aa}hle, Joseph and others},
  journal={cell},
  volume={182},
  number={2},
  pages={497--514},
  year={2020},
  publisher={Elsevier}
}

@article{andersson2020spatial,
  title={{Spatial deconvolution of HER2-positive breast tumors reveals novel intercellular relationships}},
  author={Andersson, Alma and Larsson, Ludvig and Stenbeck, Linnea and Salm{\'e}n, Fredrik and Ehinger, Anna and Wu, Sunny and Al-Eryani, Ghamdan and Roden, Daniel and Swarbrick, Alex and Borg, {\AA}ke and others},
  journal={bioRxiv},
  pages={2020--07},
  year={2020},
  publisher={Cold Spring Harbor Laboratory}
}

@inproceedings{ganguly2025merge,
  title={{MERGE: Multi-faceted Hierarchical Graph-based GNN for Gene Expression Prediction from Whole Slide Histopathology Images}},
  author={Ganguly, Aniruddha and others},
  booktitle={CVPR},
  pages={15611--15620},
  year={2025}
}

@article{jia2023thitogene,
  title={{THItoGene: a deep learning method for predicting spatial transcriptomics from histological images}},
  author={Jia, Yuran and Liu, Junliang and Chen, Li and Zhao, Tianyi and Wang, Yadong},
  journal={Briefings in Bioinformatics},
  volume={25},
  number={1},
  year={2023},
  publisher={Oxford Academic}
}

@inproceedings{dang2025hage,
  title={HAGE: Hierarchical Alignment Gene-Enhanced Pathology Representation Learning with Spatial Transcriptomics},
  author={Dang, Thao M and Li, Haiqing and Guo, Yuzhi and Ma, Hehuan and Jiang, Feng and Miao, Yuwei and Zhou, Qifeng and Gao, Jean and Huang, Junzhou},
  booktitle={MICCAI},
  pages={228--238},
  year={2025},
  organization={Springer}
}

@article{haviv2025covariance,
  title={The covariance environment defines cellular niches for spatial inference},
  author={Haviv and others},
  journal={Nature Biotechnology},
  volume={43},
  number={2},
  pages={269--280},
  year={2025},
  publisher={Nature Publishing Group US New York}
}
}

\clearpage
\setcounter{page}{1}
\maketitlesupplementary


%








\section{Additional discussion on the motivation.}
The main text briefly describes the motivation for using single-cell RNA sequencing (scRNA-seq) as a reference modality for spatial transcriptomics (ST) integration. Here, we provide additional background and discussion to complement that description.

Recent advances in high-resolution spatial transcriptomics technologies, such as Xenium, have enabled spatial profiling at near single-cell resolution. However, despite these technological developments, scRNA-seq remains a more robust and scalable reference modality due to its whole-transcriptome coverage and the widespread availability of large-scale datasets \cite{haviv2025covariance}.

Existing cell-resolved ST platforms are typically constrained by targeted gene panels and high sparsity. In many cases, only a few hundred genes are measured per cell (e.g., around 300 genes), whereas scRNA-seq experiments commonly capture several thousand genes per cell (often around 5,000). As a result, scRNA-seq generally provides higher RNA capture efficiency and more comprehensive transcriptomic coverage.

Because of these limitations, many practical analyses in spatial transcriptomics are performed at the slide or patch level rather than strictly at the single-cell resolution. These settings remain among the most common analysis scenarios in current studies.

In this context, leveraging scRNA-seq as a prior provides a promising direction for scalable integration across transcriptomic modalities. By utilizing the rich transcriptomic information available from scRNA-seq, it becomes possible to guide the analysis of spatial data and facilitate the integration of bulk RNA-seq and spatial transcriptomics data in a scalable and robust manner.

\section{Derivation of the Negative Binomial Negative Log-Likelihood}
\label{sec:sup_NB_explanation}
In the main text, we formulate the training objective as the sum of the negative log-likelihood of the assumed data distribution ($L_{\mathrm{NB}}$) and a regularization term.
In this section, we derive the negative log-likelihood function associated with the negative binomial model $L_{\mathrm{NB}}$.
The motivation of using of negative binomial distribution is to deal with over dispersion problem with count data similar to \cite{kleshchevnikov2022cell2location}.
We assume that a set of observed bulk \(\mathbf{E}^b = \{ \mathbf{e}^{\mathrm{b}(n)} \}^{N_{\mathrm{m}}}_{n=1}\), where $e^{b(n)}_g$ denotest each observed bulk gene expression count for gene $g$ in slide $n$. The set of predicted mean \(\bar{\boldsymbol{\mu}}^b = \left\{\bar{\mu}^b \left(\mathcal{X}^{(n)} \right) \right\}^{N_\mathrm{m}}_{n=1} \) and dispersion \(\boldsymbol{\theta}^b \in \mathbb{R}^{N_{\mathrm{ge}}} \).
The negative log-likelihood over a mini-batch of $N_m$ slides is therefore written as
\begin{align}
L_{\mathrm{NB}}\left(\mathbf{E}^b, \bar{\boldsymbol{\mu}}^b, \boldsymbol{\theta}^b \right)
= - \frac{1}{N_m} 
\sum_{n=1}^{N_m} 
\sum_{g=1}^{N_g} 
\log \Pr(e^{b (n)}_g \mid \mathcal{X}^{(n)}).
\end{align}

Using the probability mass function of the negative binomial distribution,
\begin{align}
\Pr(k \mid \mu, \theta)
= \frac{\Gamma(k+\theta)}{\Gamma(\theta)\,\Gamma(k+1)}
\left(\frac{\mu}{\mu+\theta}\right)^{k}
\left(\frac{\theta}{\mu+\theta}\right)^{\theta},
\end{align}
the log-probability decomposes into a sum of Gamma-function terms and logarithmic terms involving the mean and dispersion.  
Substituting this expression into the definition of $L_{\mathrm{NB}}$ yields

\begin{align}
    L_{\mathrm{NB}} =& - \frac{1}{N_m} \sum_{n=1}^{N_m} \sum_{i=1}^{N_g}  \Bigg[ \log \Gamma(e^{\mathrm{b} (n)}_g + \theta^\mathrm{b}_g) - \log \Gamma(\theta^\mathrm{b}_g) \notag \\
        & - \log \Gamma (e^{\mathrm{b}(n)}_g + 1) + e^{\mathrm{b}(n)}_g \log \left(\frac{\bar{\mu}^\mathrm{b}_g(\mathcal{X}^{(n)})}{\theta^\mathrm{b}_g + \bar{\mu}^\mathrm{b}_g(\mathcal{X}^{(n)})} \right) \notag \\
        &
        + \theta^\mathrm{b}_g \log \left(\frac{\theta_g^\mathrm{b}}{\theta^\mathrm{b}_g + \bar{\mu}^\mathrm{b}_g(\mathcal{X}^{(n)})} \right)  \Bigg].
\end{align}
Here, $\Gamma(\cdot)$ denotes the Gamma function, and the dispersion parameter $\theta^b_g$ controls the degree of overdispersion relative to a Poisson model.  
The first three terms arise from the combinatorial normalization constant of the negative binomial distribution, while the last two terms correspond to the contributions from the mean-dependent success and failure probabilities.  

\section{Processing Detail of Dataset}
\label{sec:sup_dataset}
To assess the effectiveness of the proposed method, we validated our approach using three publicly available datasets from TCGA: BRCA, KIRC, and LUAD. 
To collect the bulk dataset, we first matched gene expression data with corresponding pathology images based on sample IDs. Next, we filtered out low-quality samples in which mitochondrial gene expression exceeded 30$\%$, as these samples may indicate cellular stress. Additionally, we removed samples exhibiting abnormally high expression levels exceeding 40,000 counts. We included only samples expressing more than 5,000 genes and genes detected in at least 50 samples.
To obtain the single-cell dataset, we prepared sufficiently large datasets for each organ. Filtering was performed similarly to the bulk dataset, selecting only cells in which mitochondrial gene expression was below 30 \%. Additionally, samples with gene expression exceeding 2,500 were filtered out. We used only cells that expressed at least 200 genes and genes that were expressed in at least 200 cells.



\noindent
{\bf BRCA:}We collected 1,575 whole-slide images (WSI) and gene expression pairs from primary cases in the TCGA-BRCA cohort, encompassing 51,752 genes. After filtering, we obtained 1,474 paired samples encompassing 36,739 genes.
We used processed GSE176078\cite{wu2021single} which contains 100,064 cells encompassing 29,733 genes. 77839 cells encompassing 15621 genes are extracted by filtering.
We used the 14,047 genes common to both bulk and single-cell as the target genes for estimation. 
The canonical lineage markers were used for cell-type annotation, and 49 cell-type labels were assigned for each cell.

\noindent
{\bf KIRC:}
We collected 1,081 whole-slide images (WSI) and gene expression pairs from primary cases in the TCGA-KIRC cohort encompassing 49,873 genes. After filtering, we obtained 681 paired samples encompassing 30,723 genes.
We used RCC dataset \cite{li2022mapping} which contains 270,855 cells encompassing 19736 genes. 265825 cells encompassing 16395 genes are extracted by filtering.
We used the 14,300 genes common to both bulk and single-cell as the target genes for estimation. Annotation is added based on clustering, and 106 cell-type labels are assigned.

\noindent
{\bf LUAD:}
We collected 809 whole-slide image (WSI) and gene expression pairs from primary cases in the TCGA-LUAD cohort encompassing 49,093 genes. After filtering, we obtained 756 paired samples encompassing 31,036 genes.
We used dataset \cite{kim2020single} which contains 208,506 cells encompassing 29,634 genes. 171,128 cells encompassing 16,619 genes are extracted by filtering.
We used the 14,520 genes common to both bulk and single-cell as the target genes for estimation.



\section{Ablation Study on Prototype Construction}
To evaluate the importance of cell-type prototypes, we compare our approach with a non–cell-type baseline constructed by clustering raw gene expression profiles. 
Due to strong batch effects in the data, this baseline fails to recover meaningful cell-type partitions.

We generate prototypes using $10$, $40$, and $70$ clusters and evaluate their performance (Table~\ref{tab:ablation2}). 
The resulting SCC scores are $0.297~(-0.041)$, $0.300~(-0.038)$, and $0.303~(-0.035)$, respectively, showing no improvement over random initialization.

We further compare our method with a simple prototype construction based on averaging gene expression profiles (Average prototype). 
This averaging approach achieves an SCC of $0.329~(-0.009)$, which remains inferior to the proposed prototype formulation.

We also examine a variant without the $\alpha$ and $\beta$ parameters. 
Although a similar optimization can still be achieved (SCC: $0.331~(-0.007)$), these parameters help disentangle global scale and bias effects from cell-type–specific variations. 
This contributes not only to improved performance but also to better interpretability of the model. 
In practice, although this assessment is qualitative, the estimated cell-type proportions obtained without MC appear degraded compared with those obtained with MC.

\begin{table}[h]
\centering
\caption{Comparison of prototype construction strategies. SCC values are reported, with performance differences from the proposed method shown in parentheses.}
\label{tab:ablation2}
\begin{tabular}{lcc}
\hline
Method & Setting & SCC \\ 
\hline
Baseline & 10 clusters & 0.297 (-0.041) \\
Baseline & 40 clusters & 0.300 (-0.038) \\
Baseline & 70 clusters & 0.303 (-0.035) \\
Average prototype & -- & 0.329 (-0.009) \\
w/o $\alpha$ and $\beta$ & -- & 0.331 (-0.007) \\
Ours & cell-type informed & \textbf{0.338} \\
\hline
\end{tabular}
\end{table}

\section{Examples of cell-type proportion estimation}
\label{sec:sup_example}
\begin{figure}[t]
    \centering
    \includegraphics[width=\linewidth]{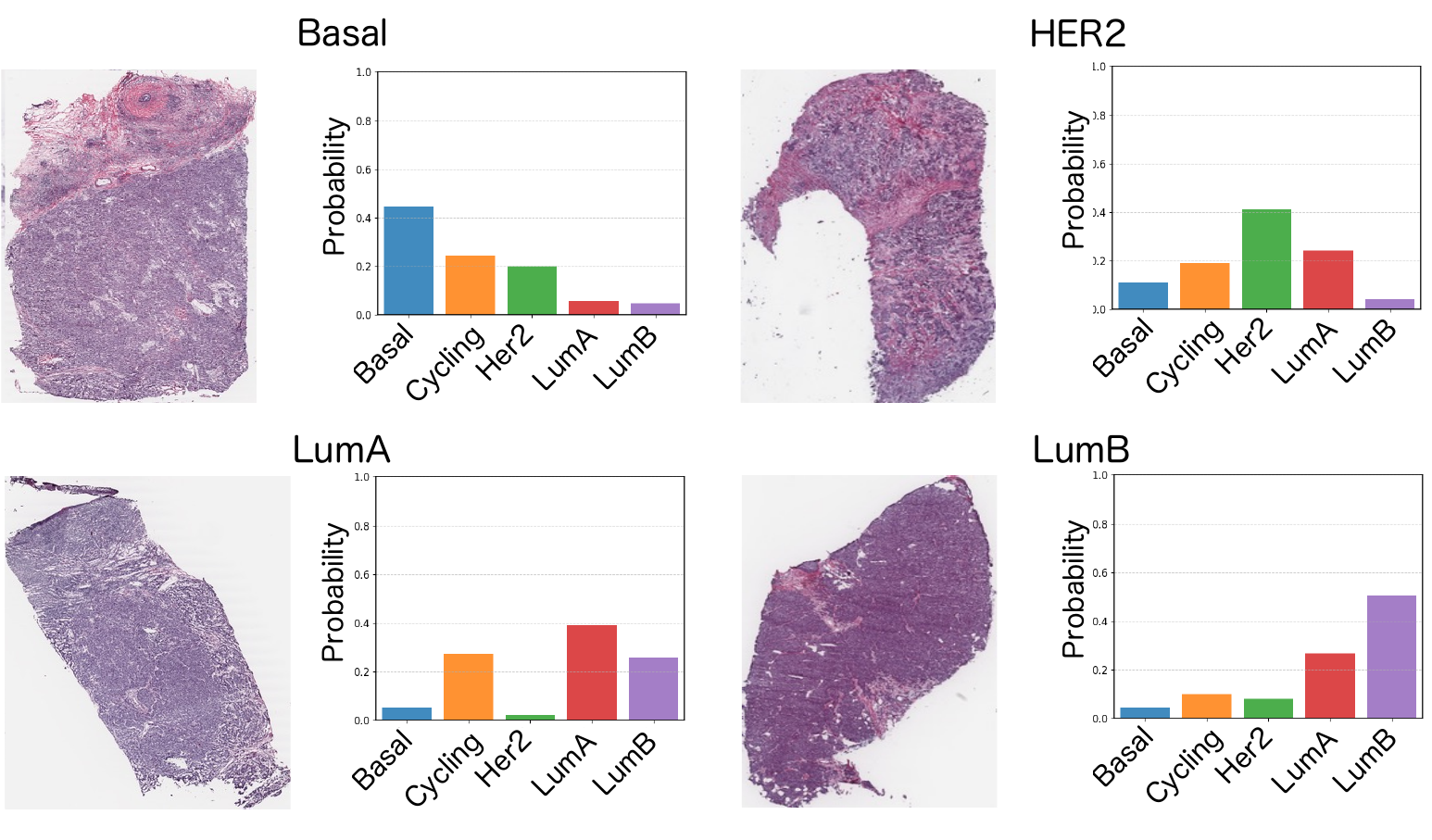}
    \caption{Example of estimated proportion.}
    \label{fig:example_proportion}
    \vspace{-5mm}
\end{figure}

We utilized BRCA subtype annotations from the cBioPortal dataset \cite{icgc2020pan} and confirmed example of estimation for each subtypes.
Leveraging single-cell annotations, the cancer-related compositional weights can be intuitively visualized.
Figure \ref{fig:example_proportion} shows examples of the estimated cell-type proportions for cancer-related subtype. The results indicate that subtype-specific cell types are accurately identified with high proportions, demonstrating that the prototype effectively leverages subtype-specific gene expression.

Because gene expression profiles are inherently high-dimensional, it is difficult to directly validate predictions using only the estimated expression values.
The proposed interpretability therefore provides a valuable means to assess the estimation validity and holds promise for improving the reliability of predictions through comparison with pathologists’ observations.
Furthermore, integrating our framework with cell segmentation could enhance both interpretability and predictive performance by modeling the spatial distribution of tumor subtypes, and extending it to jointly incorporate slide- and patch-level estimations may enable spatially resolved interpretability in gene expression prediction.


\section{Discussion about Cell-level Prototype}
\label{sec:sup_discussion}

If the correct cell-level prototype can be accurately derived from cell-level gene expression data, keeping the prototype fixed would enhance interpretability and preserve gene-gene covariance relationships. 
However, as demonstrated in our results with fixed prototypes of w/o U in Table 2, factors such as the modality gap, experimental condition gap, and observational noise make it difficult to use prototypes without updates. 

\begin{figure*}
    \centering
    \includegraphics[width=\linewidth]{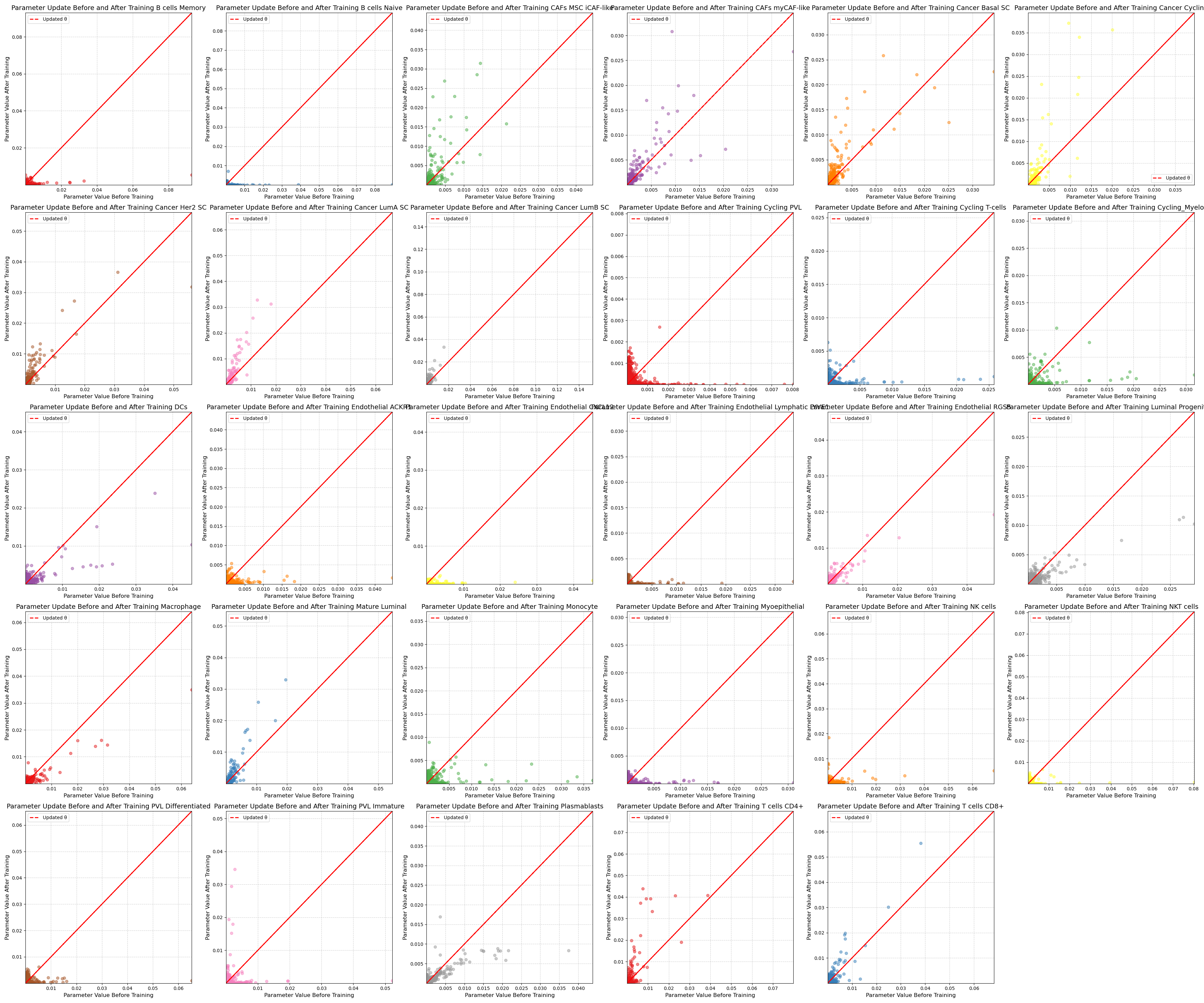}
    \caption{The parameter distribution for cell-type prototype on BRCA before and after training. Each scatter plot corresponds to a specific prototype, with individual points representing genes. The horizontal axis denotes parameter values before training, while the vertical axis represents values after training.}
    \label{fig:prototype}
\end{figure*}

Figure \ref{fig:prototype} shows the parameter distribution of the prototype on BRCA. Each subfigure indicates each cell-type prototype. As shown in Figure  \ref{fig:prototype}, the proposed method adaptively adjusts prototypes based on cell type. 
Since some cell types are easier to identify from images than others, our approach leverages data-driven prototype updates to improve gene expression estimation.

\begin{table}[t]
    \centering
    \begin{tabular}{ccc} 
        \toprule
        BRCA & KIRC & LUAD\\  \midrule
        0.338& 0.370&0.253 \\
        \bottomrule
    \end{tabular}
    \caption{Deconvolution result of cell2location.}
    \label{tab:deconv}
\end{table}

\end{document}